%% file: conference_101719.tex
\documentclass[conference]{IEEEtran}
\IEEEoverridecommandlockouts
\renewcommand{\IEEEkeywordsname}{Keywords}
\usepackage{cite}
\usepackage{amsmath,amssymb,amsfonts}
\usepackage{algorithmic}
\usepackage{graphicx}
\usepackage{textcomp}
\usepackage{xcolor}
\usepackage{url}
\usepackage{makecell}
\usepackage{enumitem}
\usepackage{multirow} % Required for multirow cells
\usepackage{adjustbox} % Required for adjusting table width
\usepackage{pgfplots}
\usepackage{arydshln} % Required for dashed lines
\pgfplotsset{compat=1.18}
\usepackage{pgfplotstable}
\usepackage{siunitx} % Required for number formatting
\usepackage[mathlines,switch]{lineno} 
\usetikzlibrary{pgfplots.groupplots}
\usepackage{tikz}

\newcommand\copyrighttext{%
  \footnotesize \textcopyright 2025 IEEE.  Personal use of this material is permitted.  Permission from IEEE must be obtained for all other uses, in any current or future media, including reprinting/republishing this material for advertising or promotional purposes, creating new collective works, for resale or redistribution to servers or lists, or reuse of any copyrighted component of this work in other works.}
\newcommand\copyrightnotice{%
\begin{tikzpicture}[remember picture,overlay]
\node[anchor=south,yshift=10pt] at (current page.south) 
  {\fbox{\parbox{\dimexpr\textwidth-\fboxsep-\fboxrule\relax}{\copyrighttext}}};
\end{tikzpicture}%
}

\def\BibTeX{{\rm B\kern-.05em{\sc i\kern-.025em b}\kern-.08em
    T\kern-.1667em\lower.7ex\hbox{E}\kern-.125emX}}
\begin{document}
\title{On-Device Adaptive Battery Power Prediction for Electric Vehicles\\
\thanks{ This research was funded by the German Federal Ministry of Research, Technology and Space (BMFTR) as part of the project KI4BoardNet under Grant 16ME0778.}
}
%Enhancing Electric Vehicle Battery Power Forecasting through On-Device Learning
% \author{\IEEEauthorblockN{Avik Bhatnagar}
% \IEEEauthorblockA{
% \textit{FZI Research Center for Information Technology}\\
% Germany \\
% bhatnagar@fzi.de}
% \and
% \IEEEauthorblockN{Sebastian Reiter}
% \IEEEauthorblockA{
% \textit{FZI Research Center for Information Technology}\\
% Germany \\
% reiter@fzi.de}
% \and
% \IEEEauthorblockN{Oliver Bringmann}
% \IEEEauthorblockA{
% \textit{University of Tübingen}\\
% Germany \\
% oliver.bringman@uni-tuebingen.de}
% }
% commented the below section for peer review
\author{\IEEEauthorblockN{Avik Bhatnagar, Anton Paule, Tobias Schuermann, Sebastian Reiter, Oliver Bringmann}
\IEEEauthorblockA{
\textit{FZI Research Center for Information Technology, University of Tübingen}\\
Germany \\
bhatnagar@fzi.de, paule@fzi.de, schuermann@fzi.de, reiter@fzi.de, oliver.bringmann@uni-tuebingen.de}
}

\maketitle
\copyrightnotice
\begin{abstract}
%Accurate battery power prediction is crucial for enhancing the performance of Electric Vehicles (EVs).
Adaptive power management in Electric Vehicles (EVs) requires accurate power prediction. Although deep learning models have emerged as highly effective for time-series forecasting in this domain, their performance is prone to degradation when exposed to data with distributions different from the training data. We introduce a novel approach that enables on-device learning in resource-constrained EV systems to continuously adapt pretrained battery prediction models to new, unseen data. We leverage existing pretrained models by transforming them into adaptable versions that retain critical hyperparameter knowledge from their initial training. We comprehensively investigate both online and offline model adaptation strategies. Our results demonstrate significant improvements in forecasting performance across various models and time horizons, achieving mean absolute error reductions of up to 7.49\% and 14.88\% with online and offline adaptation techniques, respectively. This study highlights the substantial benefit of on-device adaptation, resulting in enhanced battery power predictions than unadapted model deployments in real-world EV scenarios.
\end{abstract}


\input{Content/Introduction}
\input{Content/Methodology}
\input{Content/Experimental_Setup}
\input{Content/Results_and_discussion}
\section{Conclusion}
\label{sec:Conclusion}
%This work showcases how on-device adaptation can significantly improve model performance in the presence of data distribution shifts, specifically addressing battery power prediction in EVs. 
We demonstrated that on-device adaptation significantly improves EV battery power prediction model performance in the presence of data distribution shifts. We identified the optimal state-of-the-art time-series forecasting models for horizons (\qty{1}, \qty{2}, and \qty{3}{s}) and observed significant performance degradation under unseen summer and winter driving conditions. To mitigate this, we deployed and evaluated on-device learning using online and offline adaptation strategies for continuous model updates. To account for the computational limitations of EVs, this approach was evaluated on resource-constrained Rocket and ARM processors, examining both inference and training performance of the model. Our results demonstrate a substantial percentage error reduction of up to \qty{7.49}{\%} in an online adaptation scenario and over \qty{14.88}{\%} with offline adaptation, showcasing the effectiveness of our approach in enhancing prediction accuracy and reliability. While this approach reduces MAE, its real-world utility requires integration and thorough evaluation within a complete battery management system. Future work should focus on preventing occasional performance degradation in online adaptations and exploring its applicability to adapt to individual driving styles.

%Furthermore, the influence of hyperparameter search strategy on the number of model parameters for varying forecasting horizons could be analyzed.
%Although this approach primarily demonstrates benefits in mean absolute error reduction, its full real-world potential necessitates integration and thorough evaluation with a complete battery management system. %Future work should focus on methods to limit and prevent the occasional prediction performance degradation observed in online model adaptations. Additionally, exploring the applicability of this approach to individual driver adaptation for personalized experiences requires further investigation.

\bibliographystyle{ieeetr} 
\bibliography{reference}
\end{document}

%% file: Content/Introduction.tex
\section{Introduction}
\label{sec:Introduction}
Time-series signals, which are ubiquitous in daily life, are data recorded at regular intervals. Examples include financial market data, environmental measurements, and medical sensor data. Analyzing historical values to predict future trends offers significant benefits for proactive planning, operational optimization, and risk management. Advancements in deep learning have popularized its techniques for time-series forecasting applications \cite{gasparin_deep_2019}. Architectures such as Multi-layer Perceptrons (MLP), Convolutional Neural Networks (CNN), and Recurrent Neural Networks (RNN) demonstrate superior accuracy in future value prediction compared to conventional statistical methods \cite{torres_deep_2021}. These neural networks implicitly learn temporal features from input signals and predict single or multiple future time steps from the previous time steps \cite{lim_time_2021}.

Deep learning techniques have also been successfully applied in time series forecasting for Electric Vehicles (EVs), including predicting charging demand \cite{yi_electric_2022,zhu_electric_2019,koohfar_prediction_2023}, estimating the battery State of Charge (SOC) and residual capacity \cite{tian_state--charge_2021, de_la_iglesia_advanced_2025}. Our study extends the application of deep learning to the instantaneous battery power prediction. While existing studies cover forecasting granularities from minutes to months, we predicted future battery power consumption values for a very short-term horizon (1–3 seconds) using deep-learning models. This proactive prediction can support Battery Management System (BMS) operations, optimize the vehicle battery power supply, and prevent sudden power surge or drop peaks through load management schemes \cite{schurmann_mitigating_2022}.

However, practical battery power prediction faces challenges due to dynamically changing data distributions. Environmental factors (e.g., weather and elevation) and user behavior significantly influence the electrical power demand of EVs \cite{larusdottir_effect_2015}. Driver actions, particularly driving style, directly impact the overall electrical power requirements for the propulsion and auxiliary systems. % For example, aggressive acceleration increases the battery current demand, whereas rapid braking reduces the regenerative braking energy recovery.
Aggressive acceleration increases battery current demand, while rapid braking reduces regenerative braking energy recovery. Furthermore, electrical suspension system is affected, as aggressive maneuvers require more frequent adjustments, which consume additional energy. In addition, features such as climate control, heated seats, and phone chargers contribute to energy consumption and offer a means of balancing energy demand. Given the significant influence of environmental factors and user behavior on EVs’ electrical power demand, our approach employs deep-learning models capable of adapting to these dynamic conditions. By implementing real-time inference and continuous learning on resource-constrained in-vehicle embedded systems, our method offers enhanced data protection and reduced infrastructure costs compared to cloud-based solutions.
%This necessitates local data processing on in-vehicle embedded systems to ensure real-time power consumption estimates. It is crucial to implement both model inference and continuous learning algorithms in resource-constrained systems. %Furthermore, On-device learning for deep learning models offers several advantages: it enhances data protection and information security compared with cloud-based approaches and reduces infrastructure costs by eliminating the need for extensive sensor data transfer to the cloud.
%On-device learning for deep learning models enhances data protection and security compared with cloud-based approaches and reduces infrastructure costs by eliminating extensive sensor data transfer to the cloud.

\begin{figure*}[htpb]
    \centering
    \includegraphics[width=\textwidth]{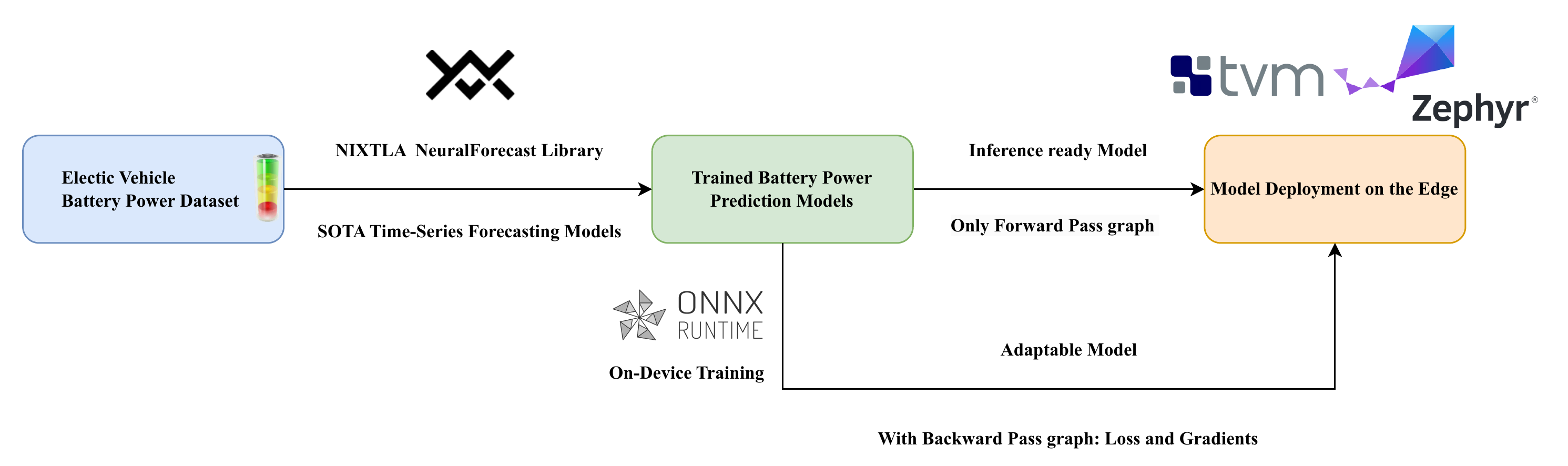}
    \caption{Workflow illustrating the training, deployment, and on-device adaptation of EV battery power prediction models.}
    %\caption{Overview Adaptable Battery Power Prediction Model Edge Deployment.}
    \label{fig:Dep}
\end{figure*}

In recent years, there has been a notable increase in research efforts towards continual learning for time-series forecasting models, with existing research predominantly employing replay memory-based techniques to store crucial learned patterns. This enables the models to continuously adapt to recent data without forgetting previously acquired knowledge, thereby minimizing forecasting errors. For instance, FSNet \cite{pham_learning_2022} updates itself with new data by augmenting each layer in the model backbone with adapters and memory units. SOLID \cite{chen_calibration_2024} uses test samples alongside contextually similar training samples to update the model's prediction layer. Another approach, OneNet \cite{zhang_onenet_2023}, employs an ensemble of forecasting networks that adapt by updating both the ensemble weights and the model parameters. However, compared with a single-forecasting model, OneNet's higher computational demands make it less suitable for edge forecasting systems.

These existing frameworks are not designed for embedded systems and do not consider the resource constraints inherent to these systems. Memory-based techniques, for instance, can lead to a substantial memory footprint when important patterns are stored in memory. Furthermore, for short prediction horizons, real-time model parameter updates on multiple data points (both current and stored) are unviable because of the computational latency observed in resource-constrained devices. Therefore, rather than relying on specialized model architectures for adaptation, this study aims to empower existing model architectures with on-device learning for continuous adaptation to new data. This paper makes the following contributions:
\begin{itemize}
\item We trained and evaluated state-of-the-art models for battery power prediction with forecasting horizons of one, two, and three seconds.
\item We demonstrate that a "one model fits all" approach is ineffective when encountering a data distribution different from the training dataset, caused by dynamic driving conditions. To address this, we enabled on-device learning through two strategies:% To address this, we enabled on-device learning in our models, facilitating adaptation through two strategies:
    \begin{itemize}[nosep]
    \item \textit{Online learning:} Model processes incoming data streams, adapts to changes in real-time. %The forecasting model continuously processes the incoming data streams and adapts to changes in real-time.
    \item \textit{Offline learning:} Model analyzes historical trips, learns power consumption patterns by iterating over data. %The forecasting model analyzes the historical trip and learns the power consumption patterns by iterating over the data multiple times.
    \end{itemize}
\item Model inference and training are executed on two resource-constrained edge devices, mimicking the limited computational capabilities of in-vehicle processors.%onboard vehicular processors.
%\item We executed both model inference and training on two distinct resource-constrained edge devices, mimicking the limited computational capabilities of onboard vehicular processors.
%\item We tested our approach by running both the model's inference and training phases on two different edge devices with limited computing power, similar to those found in vehicles.
\end{itemize}

To the best of our knowledge, this study is the first to enable deep learning-based time-series forecasting models with on-device learning capabilities on resource-constrained edge devices, allowing the models to adapt to changing operating conditions for electric vehicle power predictions. The demonstrated methodology can be seamlessly transferred to other time series forecasting domains as well.

The remainder of this paper is organized as follows. Section \ref{sec:Methodology} discusses the methodology for training state-of-the-art models for battery power prediction and the deployment of inference- and on-device learning-enabled models on resource-constrained hardware. Sections \ref{sec:Experimental Setup} and \ref{sec:Results} detail the experimental setup and discuss the results, respectively. Finally, Section \ref{sec:Conclusion} presents the conclusions and outlines future work.

%% file: Content/Methodology.tex
\section{Methodology}
\label{sec:Methodology}

\begin{figure*}[htbp]
    \centering
    \includegraphics[width=\textwidth]{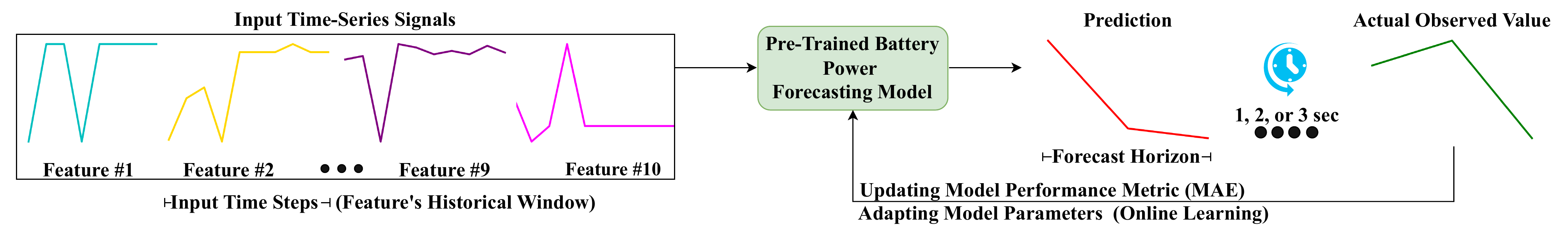}
    \caption{Schematic diagram showing online learning process.}
    \label{fig:online learning}
\end{figure*}

\subsection{State-of-the-Art Models for Battery Power Prediction}

We employed the NIXTLA NeuralForecast library \cite{olivares2022library_neuralforecast} for time-series forecasting, leveraging its implementation of various state-of-the-art models. We utilized historical exogenous features and past battery power values for battery power forecasting to generate univariate forecasts. We trained deep learning models to implicitly learn from these multiple input time series and to predict the output power. The models considered included Multi-Layer Perceptron (MLP) based architectures, such as Neural Basis Expansion Analysis with exogenous variables (N-BEATSx) \cite{olivares_neural_2023}, Deep Non-Parametric Time Series forecaster (DeepNPTS) \cite{rangapuram_deep_2023}, Time-series Dense Encoder (TiDE) \cite{das_long-term_2024}, and Neural Hierarchical Interpolation for Time Series Forecasting (N-HiTS) \cite{challu_n-hits_2022}. Additionally, we evaluated the 1D convolutional neural network Bidirectional Temporal Convolutional Network (BiTCN) \cite{sprangers_parameter-efficient_2023}, the recurrent neural network-based Temporal Fusion Transformer (TFT)\cite{lim_temporal_2021}, and Kolmogorov–Arnold Networks (KAN)\cite{liu2025kankolmogorovarnoldnetworks}.%, an alternative to MLP models.

We trained these models with prediction horizons of \qty{1},\qty{2}, and \qty{3}{s} using a direct forecasting strategy, in which the model simultaneously makes all predictions rather than employing a regressive, sequential approach. Given the non-trivial nature of multi-time step forecasting, we optimized the hyperparameters using the Ray tuning library \cite{liaw2018tune} with default search spaces for each model and forecasting horizon. This approach identified optimally trained models for each forecasting horizon, accounting for variations in architecture, input time steps, learning rate, batch size, and number of layers.

\subsection{Edge AI Deployment and Continuous Adaptation}

An overview of the edge AI deployment and training for the battery power prediction models can be seen in Fig. \ref{fig:Dep}. The TVM machine learning compiler \cite{chen2018tvmautomatedendtoendoptimizing} facilitates the deployment of machine learning models on diverse hardware backends. It optimizes model deployment for target hardware through high-level graph and kernel optimizations. Integrating the compiled model with real-time operating systems, such as Zephyr \cite{zephyrproject}, enables model inference on resource-constrained embedded devices \cite{peccia2024efficientedgeaideploying}. This approach enables the deployment of trained prediction models, allowing on-device predictions from new data. Beyond inference, we enable on-device learning by statically appending the forward computation graph with a backward pass at compile time. We obtain this training graph from the artifact generation phase of the ONNX Runtime's on-device training support \cite{onnxruntime}. This enables backpropagation, where we compute the gradients of the Mean Absolute Error (MAE) loss function with respect to the model parameters for continuous adaptation. The resultant graph can be executed on-device using the same framework as the inference-only model deployment. We extended TVM frontend support for operators specific to loss and gradient calculation to accommodate the new ONNX training graph. 

We illustrate our online continuous adaptation strategy for the model updates in Fig. \ref{fig:online learning}. Unlike the initial batch training, this approach continuously updates the model using the latest input data. The model generates a prediction, after prediction horizon passes and the actual value becomes available, we update the MAE. Using this loss, the gradients are calculated, and the model parameters are updated. This methodology ensures that the model performance is always evaluated on data points before they are used for training, effectively preventing data leakage. We employed a lower learning rate for the online strategy. As the model is updated continuously on single data points, this lower learning rate prevents aggressive updates based on outliers or noise, thereby safeguarding the long-term model performance. To update the models online at iteration $t$, we apply the general update rule based on the stochastic gradient descent:
\begin{equation}
    \theta_{m, t+1} = \theta_{m, t} - \eta_{\text{online}, m} \cdot \nabla L(f_m(\theta_{m, t}, x_i), y_i)
    \label{eq:single_point_update}
\end{equation}

Here, $f_m$ denotes the output for model $m$, and $L(f_m(\theta_{m, t}, x_i), y_i)$ is the loss evaluated on the prediction of model $m$ for input $x_i$ against the true label $y_i$. $\theta_{m, t+1}$ represents the updated model parameters, obtained using the online learning rate $\eta_{\text{online}, m}$, and $\nabla L()$ is the computed gradient of the loss function with respect to the model parameters $\theta_{m, t}$.

An alternative strategy for on-device learning, similar to initial model training, involves batch gradient descent. This approach utilizes the training graph to iterate through the stored data for multiple epochs, updating the model parameters in batches. However, this method is only feasible in an offline manner because it requires the entire dataset for its multi-epoch and multi-batch training strategy. Here, the update rule  based on mini-batch gradient descent can be expressed as:
\begin{equation}
    \theta_{m, t+1} = \theta_{m, t} - \eta_{\text{offline}, m} \cdot \frac{1}{B_m} \sum_{j=1}^{B_m} \nabla L(f_m(\theta_{m, t}, x_j), y_j)
    \label{eq:batch_update}
\end{equation}

For each model $m$ at training iteration $t$, a batch of data points of the same size as that used for its training ${B_m}$ is taken, and the mean of gradients over the entire batch is used, along with the offline learning rate $\eta_{\text{offline}, m}$ to obtain the updated parameters $\theta_{m, t+1}$.

%% file: Content/Experimental_Setup.tex
\section{Experimental Setup}
\label{sec:Experimental Setup}
\textbf{Dataset:} We utilized the Battery and Heating Data in Real Driving Cycles Dataset from the Technical University of Munich, Institute of Automotive Technology \cite{6jr9-5235-20}. This dataset comprises 72 real driving trips conducted using a BMW i3 (\qty{60}{Ah}) under both summer and winter conditions. Each trip included environmental (e.g., temperature and elevation), vehicular (e.g., speed and throttle), battery-specific (e.g., voltage, current, temperature, and SOC), and heating circuit (e.g., indoor temperature and heating power) data.  We derived the instantaneous battery power by multiplying the battery voltage and current signals. The original \qty{100}{ms} sampling rate was upsampled to \qty{1}{s} to ensure a minimum 1-second forecasting horizon for time series forecasters and facilitate model deployment on resource-constrained devices where execution time may exceed \qty{100}{ms}.

For model evaluation, we reserved the randomly choosen trip numbers 32 (summer) and 37 (winter) as unseen data to independently assess seasonal performance. The remaining trips were individually split into \qty{70}{\%} training and \qty{30}{\%} testing sets. We performed a correlation analysis on the training data to identify the most significant input features for forecasting the battery power. This analysis yielded the following features: Battery Current [A], Throttle [\%], Motor Torque [Nm], Longitudinal Acceleration [m/s$^2$], Battery Voltage [V], Velocity [km/h], Regenerative Braking Signal, Requested Heating Power [W], CAN Heating Power [KW], and last observed Battery Power [W]. These input features were then standardized using the mean and standard deviation of the training dataset.

\textbf{Edge Devices:} We considered two processors with different processing capabilities to represent resource-constrained compute environments in vehicles. The first is a quad-core Arm Cortex-A53 processor operating at \qty{1.2}{GHz} with \qty{4}{GB} of RAM, present on the processing system of the AMD Zynq UltraScale+ MPSoC ZCU102 Evaluation Board. The second is a RISC-V based Rocket Processor Core, which we implemented on the programmable logic side of the same FPGA board. This Rocket Core operates at \qty{150}{MHz}, with \qty{16}{KB} L1 (Instruction and Data) and \qty{2}{MB} L2 caches, and \qty{512}{MB} of external DRAM.

\textbf{Update Strategies:} Initial model training used learning rates ranging from $2.48 \times 10^{-4}$ to $9.53 \times 10^{-2}$ and batch sizes 64 to 256. For online continuous model adaptation, we selected a learning rate equal to 0.1 times the training learning rate for each model and prediction horizon. Lower learning rate is a common practice for fine-tuning models with limited data, ensuring conservative updates, and balancing learning from new data points while retaining prior prediction knowledge. This strategy allows for on-the-fly model updates, where performance improves as a trip progresses compared to unadapted models. Alternatively, an offline update strategy is used to adapt the model to recent trip data after trip completion to improve the performance of upcoming trips. This involves iterating through the trip data multiple times and updating the model with the same batch gradient settings (learning rate and batch size) used in the initial training for 100 epochs. In this batch learning setting, the gradients are averaged over the entire batch, preventing overfitting to a single observation, unlike online adaptation. Consequently, the original learning rate is appropriate here, whereas it would lead to suboptimal adaptation and degraded long-term performance in the online case. Offline adaptation is also computationally more intensive than online adaptation because it processes data for multiple epochs, whereas online adaptation processes data only once.

%% file: Content/Results_and_discussion.tex
\section{Results and Discussion}
\label{sec:Results}

This section presents a comprehensive analysis of our experimental results, focusing on the model performance across various forecasting horizons and selected processors, the impact of data distribution shifts, and the effectiveness of on-device adaptation strategies. Our findings highlight the challenges posed by evolving data environments and demonstrate the benefits of both offline and online model adaptation for mitigating performance degradation.

\begin{figure}[htbp]
    \centering
    \includegraphics[width=\columnwidth]{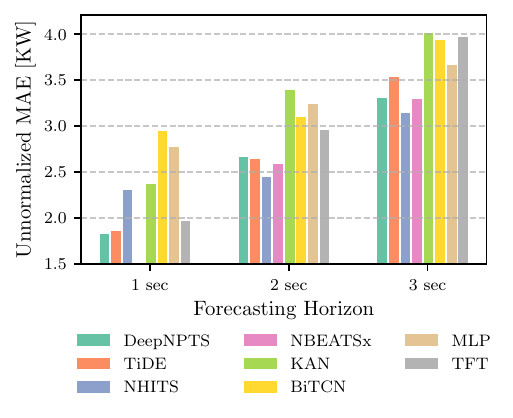}
    \caption{Performance of forecasting models on the test set across time horizons.}
    \label{fig:SOTA_MAE}
\end{figure}

Fig. \ref{fig:SOTA_MAE} illustrates the unnormalized MAE values for the trained models across various forecasting horizons on the test dataset. A clear trend emerges, showing that the error increases with an increasing forecasting horizon, indicating that all models degrade in performance as the forecasting horizon lengthens. The maximum observed error at a \qty{3}{s} forecasting horizon was approximately \qty{4}{KW}, which remained below \qty{2.2}{\%} of the total range of battery power values (\qty{180}{KW}). NBEATSx was not included for the 1-second horizon due to its inherent requirement for a longer forecasting horizon to effectively model seasonal and trend components.

\begin{figure}[htbp]
    \centering
    \includegraphics[width=\columnwidth]{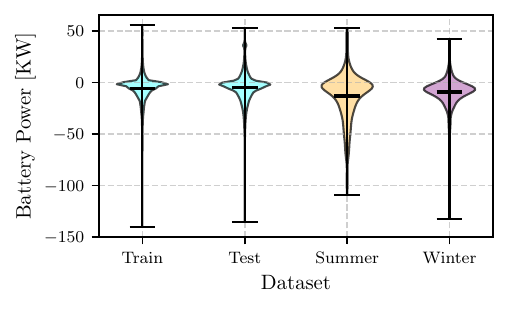}
    \caption{Data distribution of the different datasets.}
    \label{fig: Data_dist}
\end{figure}

\begin{figure}[htbp]
    \centering
    \includegraphics[width=\columnwidth]{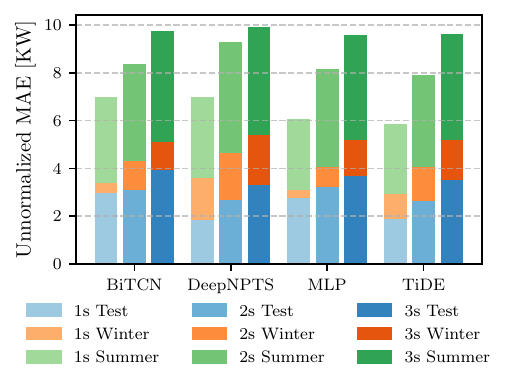}
    \caption{Model performance on test and unseen datasets.}
    \label{fig:Summer_Winter_MA}
\end{figure}

Fig. \ref{fig: Data_dist} illustrates the data distributions of the training, test, summer (trip 32), and winter (trip 37) datasets. The violin plots depict the data density, along with the min, max, and mean power values for each dataset. A clear shift in data distribution was evident between the individual summer and winter trips compared to the distribution used for model training. This shift also extends to other input features. Such shifts degrade the model performance, as demonstrated in the subsequent Fig. \ref{fig:Summer_Winter_MA}. This observation underscores the need for on-device model adaptation to improve performance for specific trips rather than relying on a single model for all scenarios.

Based on preliminary performance, we selected the BiTCN, MLP, TiDE, and DeepNPTS models for on-device adaptation, representing both lower and higher performing forecasters. As detailed in the methodology, we enabled on-device learning for these models and calculated the gradients necessary for parameter updates. For the DeepNPTS model, the batch-normalization layers were frozen during the single-data-point updates to maintain stable normalization, avoiding corruption from noisy, single-sample statistics.  We also evaluated the effect of the data distribution shift on the performance of these models. Fig. \ref{fig:Summer_Winter_MA} presents overlapped bar plots that illustrate the MAE of the trained models when evaluated on the test dataset and the unseen summer and winter trips. For all models and forecasting horizons, the observed MAE for the unseen summer and winter trips was considerably higher than that for the test dataset. The increase in error is more pronounced for the summer trips, showing an average increase of approximately \qty{184}{\%}, compared to a \qty{46}{\%} average increase for the winter trips. This behavior can be attributed to the greater extent of the data distribution shift observed in the summer trip power values, as shown in Fig. \ref{fig: Data_dist}.

\begin{table}[!t]
\centering
\caption{Model characteristics on selected processors across different horizons.}
\label{tab:model_characteristics}
\adjustbox{max width=\columnwidth}{%
\begin{tabular}{|l|l|c|c|c|c|}
\hline
\textbf{Horizon} & \textbf{Model Characteristics} & \textbf{BiTCN} & \textbf{MLP} & \textbf{TiDE} & \textbf{DeepNPTS} \\
\hline
\multirow{6}{*}{\makecell[l]{3 sec}} & Rocket Inf. (s) & 0.0063 & 0.0626 & 0.3229 & 0.0024 \\
& Rocket Train. (s) & 0.0321 & 0.3167 & 1.8322 & 0.0111 \\
& ARM Inf. (s) & 0.0003 & 0.0035 & 0.0164 & 0.0001 \\
& ARM Train. (s) & 0.0016 & 0.0121 & 0.0804 & 0.0006 \\
\cdashline{2-6}
& \# Parameters & 4,740 & 558,083 & 2,319,126 & 20,571 \\
& Param Size (MB) & 0.018 & 2.129 & 8.847 & 0.078 \\
\hline
\multirow{6}{*}{\makecell[l]{2 Sec}} & Rocket Inf. (s) & 0.0085 & 0.1204 & 0.0789 & 0.0004 \\
& Rocket Train. (s) & 0.0448 & 0.6264 & 0.3895 & 0.0018 \\
& ARM Inf. (s) & 0.0006 & 0.0068 & 0.0018 & 0.0001 \\
& ARM Train. (s) & 0.0020 & 0.0220 & 0.0079 & 0.0004 \\
\cdashline{2-6}
& \# Parameters & 13,251 & 1,072,642 & 637,784 & 3,532 \\
& Param Size (MB) & 0.050 & 4.092 & 2.433 & 0.013 \\
\hline
\multirow{6}{*}{\makecell[l]{1 sec}} & Rocket Inf. (s) & 0.0047 & 0.2778 & 0.6772 & 0.0002 \\
& Rocket Train. (s) & 0.0212 & 1.6194 & 4.1255 & 0.0011 \\
& ARM Inf. (s) & 0.0003 & 0.0156 & 0.0392 & 0.0001 \\
& ARM Train. (s) & 0.0012 & 0.0791 & 0.1985 & 0.0004 \\
\cdashline{2-6}
& \# Parameters & 8,962 & 2,131,969 & 5,359,031 & 1,845 \\
& Param Size (MB) & 0.034 & 8.133 & 20.443 & 0.007 \\
\hline
\end{tabular}%
}
\end{table}

The computation latencies of these models on the selected processor cores are presented in Table \ref{tab:model_characteristics}. Inference latency represents the time required for a model to execute and make predictions on the input data, whereas training latency includes the additional computation for loss and parameter gradients calculation. The model parameter counts vary with the forecasting horizons owing to the hyperparameter optimization for each horizon. The table also lists the parameter size required to store the model weights and biases; note that training typically requires at least double the memory for gradients. The training execution time is generally 3-6 times longer than the inference time for both processors. This is because gradient calculation during training requires more operations; for instance, a convolution layer requires both transposed convolution and convolution operations to compute gradients with respect to its input and weights. The DeepNPTS model is generally the most lightweight in terms of computational resource demand, whereas the TiDE and MLP models tend to have a higher number of parameters, resulting in higher model execution times. Generally, both model training and inference latencies are below the forecasting horizon, enabling online model updates using the previous predictions and ground truth values. However, for a 1-second forecasting horizon, the TiDE and MLP models exhibit training times of \qty{4.1}{s} and \qty{1.6}{s}, respectively, on the Rocket core. These times significantly exceed the 1-second update window due to the limited computing capability of the processor. This can be resolved by utilizing a more powerful processor, such as an ARM processor, to achieve a truly online adaptive prediction system. On the other hand, the offline adaptation strategy introduces additional overhead compared to the online strategy. In our experiments, the training latency of offline adaptation is scaled by a factor of 100 with the number of epochs. Furthermore, offline adaptation requires the entire trip's data to be stored. Also, the gradients stored can be 64 to 256 times larger, depending on the batch size for the respective model. %This also includes storage requirements for the gradients, which can be 64 to 256 times larger, depending on the batch size used for the respective model.

\begin{table}[!t]
\centering
\caption{Percentage Error Reduction - Online learning}
\label{tab:online adapt}
\resizebox{\columnwidth}{!}{%
\begin{tabular}{|l|*{3}{c|}*{3}{c|} c|}
\hline
\textbf{SOTA Model} & \multicolumn{3}{c|}{\textbf{Summer}} & \multicolumn{3}{c|}{\textbf{Winter}} & \textbf{Mean} \\
\cline{2-7}
& \textbf{1s} & \textbf{2s} & \textbf{3s} & \textbf{1s} & \textbf{2s} & \textbf{3s} & \\
\hline
DeepNPTS & -0.46 & 1.4 & -0.16 & 0.71 & -0.05 & 0.11 & 0.26 \\
MLP & -0.5 & 2.23 & 0.12 & 1.16 & 0.04 & 0.11 & 0.53 \\
TiDE & 4.95 & 1.7 & 0.95 & 2.25 & 0.61 & 0.12 & 1.76 \\
BiTCN & 7.49 & -1.24 & 0.38 & 5.16 & 0.73 & 1.48 & 2.33 \\
\hline
\end{tabular}%
}
\end{table}

\begin{table}[!t]
\centering
\caption{Percentage Error Reduction  - Offline learning}
\label{tab:offline adapt}
\resizebox{\columnwidth}{!}{%
\begin{tabular}{|l|*{3}{c|}*{3}{c|} c|}
\hline
\textbf{SOTA Model} & \multicolumn{3}{c|}{\textbf{Summer}} & \multicolumn{3}{c|}{\textbf{Winter}} & \textbf{Mean} \\
\cline{2-7}
& \textbf{1s} & \textbf{2s} & \textbf{3s} & \textbf{1s} & \textbf{2s} & \textbf{3s} & \\
\hline
DeepNPTS & 5.3 & 5.83 & 2.17 & 3.74 & 3.64 & 2.3 & 3.83 \\
MLP & 8.08 & 10.85 & 2.3 & 9.79 & 7.06 & 2.16 & 6.71 \\
TiDE & 13.82 & 6.07 & 5.5 & 10.18 & 4.52 & 4.78 & 7.48 \\
BiTCN & 14.88 & 5.13 & 9.37 & 11.09 & 7.11 & 9.29 & 9.48 \\
\hline
\end{tabular}%
}
\end{table}

Tables \ref{tab:online adapt} and \ref{tab:offline adapt} illustrate the relative error reduction achieved through offline and online model adaptations, respectively. As anticipated, the percentage reduction in error is significantly higher for offline adaptation than for online adaptation, primarily because offline adaptation allows for model updates over multiple epochs. However, it is crucial to note that offline adaptation evaluates the model on the same data on which it was trained, whereas online adaptation updates metrics based on predicted values and subsequently received ground truths. This makes online adaptation a more practical evaluation method for real-world deployments. Although a slight performance degradation, as can be seen from the negative values for the percentage error reduction, occurs occasionally with online adaptation, it remains relatively low compared to the observed improvements. This degradation, compared to the baseline unadapted model, can be attributed to the limitations of the stochastic gradient descent used for updating the model parameters, resulting in a local optimum and, therefore, a limited model performance.

As observed from the mean percentage error reduction values across the trips, the most significant model improvement for both adaptation strategies was observed in the TiDE and BiTCN models, which represent the top and bottom performers, respectively, in our initial selection. However, the performance improvement for DeepNPTS is limited because the frozen batch-normalization layers are not updated. Furthermore, for a given trip, the online adaptation performance improved as the forecasting horizon decreased. This is because the model can be updated on more data points when data are processed in a non-overlapping fashion, such as with a \qty{1}{s} forecasting horizon compared to \qty{3}{s}. 
In conclusion, these adaptation strategies enable models to effectively adapt to unseen data and evolving environments, leading to improved performance compared to unadapted models.
%In conclusion, employing these adaptation strategies enables models to adapt to unseen data and evolving environments, achieving improved performance compared to unadapted models.%an error reduction of up to 7.49\% in an online fashion and over 14.88\% in an offline fashion.